\documentclass[preprint,12pt, a4paper]{elsarticle}

\usepackage{amssymb}
\usepackage{fancyvrb}

\usepackage{lineno}
\usepackage{amsmath}
\usepackage{algorithm2e}
\usepackage{graphicx}
\usepackage{float}
\usepackage{amssymb}
\usepackage{caption}
\usepackage{subcaption}
\usepackage{multirow}
\usepackage{ctable}
\usepackage{calc}
\usepackage{mathtools}
\usepackage{url}
\usepackage{hyperref}

 \makeatletter 
 \providecommand{\doi}[1]{%
   \begingroup
     \let\bibinfo\@secondoftwo
     \urlstyle{rm}%
     \href{http://dx.doi.org/#1}{%
       doi:\discretionary{}{}{}%
       \nolinkurl{#1}%
     }%
   \endgroup
 }
 \makeatother

\newcommand{\argmax}{\arg\!\max}
\usepackage{color}

\journal{SoftwareX}

\begin{document}

\begin{frontmatter}



\title{varrank: an R package for variable ranking based on mutual information with applications to observed systemic datasets}

\author[label1]{Gilles Kratzer}
\author[label1,label2]{Reinhard Furrer}

\address[label1]{Department of Mathematics, University of Zurich, Zurich, \underline{Switzerland}}
\address[label2]{Department of Computational Science, University of Zurich, Zurich, \underline{Switzerland}}

\begin{abstract}
This article describes the R package  { \sf varrank}. It has a flexible implementation of heuristic approaches which perform variable ranking based on mutual information. The package is particularly suitable for exploring multivariate datasets requiring a holistic analysis. The core functionality is a general implementation of the minimum redundancy maximum relevance (mRMRe) model. This approach is based on information theory metrics. It is compatible with discrete and continuous data which are discretised using a large choice of possible rules. The two main problems that can be addressed by this package are the selection of the most representative variables for modeling a collection of variables of interest, i.e., dimension reduction, and variable ranking with respect to a set of variables of interest.
\end{abstract}


\end{frontmatter}



\section{Motivation and significance}\label{intro}

A common challenge encountered when working with high dimensional datasets is that of variable selection.  All relevant confounders must be taken into account to allow for unbiased estimation of model parameters, while balancing with the need for parsimony and producing interpretable models \cite{walter2009variable}. This task is known to be one of the most controversial and difficult tasks in epidemiological analysis, yet, due to practical, computational, or time constraints, it is often a required step. We believe this applies to many multivariable holistic analyses, independent of the research field.

Systems epidemiology, in an interdisciplinary effort, aims to include individual, meta population and possibly environmental information with a focus on a disease's dynamic understanding. Systems thinking, with particular emphasis on analysing multiple levels of  causation, allows epidemiologists to discriminate between directly and indirectly related contributions to a disease or set of outcomes  \cite{dammann2014systems}. One key characteristic of this approach is to balance prior knowledge of disease dynamics from previous or parallel studies with metapopulation, environmental or ecological contributions. The set of possible variable candidates is usually immense, but in practice adding all variables is often not suitable as it can decrease the global model predictive efficiency. Only a part of the model is known before collecting the data. In this context, the most widely used approach for variable selection is based on prior knowledge from the scientific literature \cite{walter2009variable}.

Thanks to its increasing popularity in epidemiology, the open source statistical software R \cite{team2017r} is a convenient environment in which to distribute implementation of new approaches. Here we present an implementation of a collection of model-free algorithms capable of working with a large collection of candidate variables based on a set of variables of importance. It is called model-free as it does not suppose any pre-specified model. Contrary to existing R packages, the new package {\sf varrank} deals with a set of variables of interest and allows the user to select from various methods and options for the optimization algorithm. It also contains a plotting function which helps in analyzing the data. Finally, it is based on an appealing approach that does not rely on goodness-of-fit metrics to measure variable importance but rather it measures relevance penalized by redundancy.  
 
\subsection{Previous research}\label{previous_work}

Variable selection approaches, also called feature or predictor selection in other contexts, can be categorized into three broad classes: filter-based methods, wrapper-based methods, and embedded methods \cite{guyon2003introduction}. They differ in how the methods combine the selection step and the model inference. Filter-based approaches perform variable selection independently of the model learning process, whereas wrapper-based and embedded methods combine these steps. An appealing filter approach based on mutual information (MI) is the minimum redundancy maximum relevance (mRMRe) algorithm \cite{battiti1994using, kwak2002input, estevez2009normalized}, which has a wide range of applications \cite{peng2005feature}. The purpose of this heuristic approach is to select the most relevant variables from a set by penalising according to the amount of redundancy variables share with previously selected variables. At each step, the variables that maximize a score are selected. The mRMRe approach is based on the estimation of information theory metrics (see \cite{cover2012elements} for classical definitions). In epidemiology, the most frequently used approaches to tackle variable selection based on modeling use goodness-of fit metrics. The paradigm is that important variables for modeling are variables that are causally connected and predictive power is a proxy for causal links. On the other hand, the mRMRe algorithm aims to measure the importance of variables based on a relevance penalized by redundancy measure which makes it appealing for epidemiological modeling.

The mRMRe approach, originally proposed by \citet{battiti1994using}, can be described as an ensemble of models \cite{van2010increasing} whereas the general term `mRMRe' has been coined by \citet{peng2005feature}. A general formulation of the ensemble of the mRMRe technique is as follows. Assume we have a global set of variables $\textbf{F}$ and a subset of important variables $\textbf{C}$. The variables in set $\textbf{C}$ are the variables the user wants in the final model as they are supposed to be important to modeling. Moreover, let $\textbf{S}$ denote the set of already selected variables and $f_i$ a candidate variable. The local score function is expressed as 

\begin{equation}\label{eq:mRMR}
        g(\alpha,\textbf{C}, \textbf{S}, f_i) = \underbrace{\text{MI}(f_i;\textbf{C})}_\textrm{Relevance}\, - \sum_{f_s \in \textbf{S}} \overbrace{\alpha(f_i, f_s,\textbf{C}, \textbf{S})}^\textrm{Scaling factor}\, \underbrace{\text{MI}(f_i; f_s)}_\textrm{Redundancy}.
\end{equation}

The list below presents four possible values of the normalizing function $\alpha$ that define four implemented models in {\sf varrank}.

\begin{enumerate}
\item $\alpha(f_i,f_s,\textbf{C}, \textbf{S})=\beta$, where $\beta>0$ is a user defined parameter. This model is called the mutual information feature selector (MIFS) in \citet{battiti1994using}.
\item $\alpha(f_i,f_s,\textbf{C}, \textbf{S})=\beta ~ {\text{MI}(f_s;\textbf{C})}/{\text{H}(f_s)}$, where $\beta>0$ is a user defined parameter. This model is called MIFS-U in \citet{kwak2002input}.
\item $\alpha(f_i,f_s,\textbf{C}, \textbf{S})={1}/{|\textbf{S}|}$, which is called min-redundancy max-relevance (mRMR) in \citet{peng2005feature}.
\item $\alpha(f_i,f_s,\textbf{C}, \textbf{S})={1}/\{{|\textbf{S}|\text{min}(\text{H}(f_i),\text{H}(f_s))\}}$, called normalized MIFS in \citet{estevez2009normalized}.
\end{enumerate}

For easier reference, the methods are called \textit{battiti}, \textit{kwak}, \textit{peng} and \textit{esteves} in the R package {\sf varrank}. The first and second terms on the right-hand side of~\eqref{eq:mRMR} are local proxies for the relevance and the redundancy of a variable $f_i$, respectively. Redundancy is used to avoid selecting variables that are highly correlated with previously selected ones. Local proxies are needed, as computing the joint MI between high dimensional vectors is computationally very expensive. There exists two popular ways to combine relevance and redundancy: either take the difference (\textit{mid}), as in \eqref{eq:mRMR}, or the quotient (\textit{miq}). This criteria can be embedded into a greedy search algorithm that locally optimizes the variable choice. In~\eqref{eq:mRMR}, the function $\alpha$ attempts to shrink left side and right side terms to the same scale. In \citet{peng2005feature} and \citet{estevez2009normalized}, the ratio of comparison is adaptively chosen as $\alpha=1/|\textbf{S}|$ in order to control the right term, which is a cumulative sum that increases quickly as the cardinality of $\textbf{S}$ increases. The function $\alpha$ normalizes the right side.

\subsection{Available R packages on CRAN for variables selection}

One popular R package for variable selection is {\sf caret} \cite{kuhn2008caret}, which uses classification and regression training to select variables. Three other popular R packages based on the random forest methodology are {\sf Boruta} \cite{kursa2010feature}, {\sf varSelRF }\cite{diaz2007genesrf} and {\sf FSelector} \cite{romanski2016package}. {\sf Boruta} has an implementation of a variable selection procedure that aims to find all variables carrying information useful to prediction. {\sf varSelRF} targets the analysis of gene expression datasets. {\sf FSelector} contains algorithms for filtering attributes and for wrapping classifiers.

Lastly, the package {\sf mRMRe} \cite{de2013mrmre} has a fast parallel implementation of the model described in \citet{peng2005feature}. It can deal with continuous, categorical, and survival variables. The mutual information is estimated through a linear approximation based on correlation. This is the closest R package to {\sf varrank}.

\section{Software description}
\label{sftdesc}

The package {\sf varrank} is implemented in R \cite{team2017r}. It contains documentation with examples and comparisons to alternative approaches and unit tests implemented using the {\sf testthat} functionality \cite{wickham2011testthat}. 

In systems epidemiology the data are typically a mix of discrete and continuous variables. Thus, a common, popular and efficient choice to compute information metrics is to discretize the continuous variables and then deal with discrete variables only. 
Some static univariate unsupervised splitting approaches that are computationally very efficient are implemented in {\sf varrank} \cite{garcia2013survey}. In the current implementation, several popular histogram-based approaches are implemented: Cencov's rule \cite{cencov1962estimation}, Freedman-Diaconis' rule \cite{freedman}, Scott's rule \cite{scott}, Sturges' rule \cite{sturges}, Doane's formula \cite{doane} and Rice's rule. 
Although not recommended, it is possible to manually select the number of bins.
The MI is estimated through the count of the empirical frequencies within a plug-in estimator. An alternative approach is to use clustering with the elbow method to determine the optimal number of clusters \cite{goutte1999clustering}. This method is implemented using a fixed ratio of the between-group variance to the total variance. Two approaches compatible only with continuous variables are also implemented, one based on correlation \cite{cover2012elements} and the other one based on nearest neighbors \cite{kraskov2004estimating}.

\subsection{Software architecture}

The workhorse of the {\sf varrank} package is the sequential forward implementation of Algorithm~\ref{algo:mRMR}. The first variable  is selected using a pure relevance metric. The following variables are selected sequentially using the local score until one reaches a count wall or only one variable remains.  The backward implementation prunes the set of candidates, see Algorithm~\ref{algo:mRMR}. The two algorithms are described in the supplementary material. 

The \textit{varrank()} function returns a list that contains two entries: the ordered list of selected variables and the matrix of the scores. This object belongs to the S3 class \textit{varrank}, enabling the use of R's ``object oriented functionalities''. Three S3 methods are currently implemented: the print method displays a condensed output, the summary method displays the full output and a plot method. The plot method is an adapted version of an existing plot function from \citet{gregory2016gplots}.

\subsection{Software functionalities}

The required input arguments for the \textit{varrank()} function are:
 
\begin{description}
\item[data.df] a data frame with columns of either numeric or factor class;
\item[variable.important] a list containing the set's names of variables of importance. This set has to be in the input data frame;
\item[method] specification of $\alpha$ in~\eqref{eq:mRMR}. This can be: \textit{battiti, kwak, peng,} or \textit{esteves}. The user defined parameter is called \textit{ratio};
\item[algo] the algorithm. This can be: \textit{forward}, or  \textit{backward} (see Algorithm~\ref{algo:mRMR}  in Appendix~\ref{appendix:algo});
\item[scheme]  the search scheme to be used. This can be \textit{mid}, or \textit{miq}, which stand for the mutual information difference and quotient schemes, respectively. Those are the two popular ways to combine the relevance and redundancy.
\item[discretization.method] the discretization method. See section \ref{sftdesc} for details.
\end{description}

Optionally the user can provide the number of variables to be returned and a logical parameter for displaying a progress bar. The function returns a list containing the variables and their scores in decreasing order for a forward search (or increasing order list for a backward search). The comprehensive matrix of scores is returned. The matrix of scores is sequentially computed with eq. \eqref{eq:mRMR}. This matrix is a triangular matrix because the scores are computed only with the remaining variables (the ones not yet selected). The maximum local scores on the diagonal are used at the selection step. Detailed help files are included in the {\sf varrank} R package.

\section{Illustrative examples}

In this section, we use three classical example datasets to illustrate the use, performance and features of {\sf varrank}. (i) The Longley dataset \cite{team2017r} contains seven continuous economical variables observed yearly from 1947 to 1962 (16 observations). This small dataset is known to have highly correlated variables. (ii) The Pima Indians Diabetes dataset \cite{leischmlbench} contains 768 observations on nine clinical variables relating to diabetes status. (iii) The EPI dataset \cite{revelle2014psych} contains 57 variables measuring two broad dimensions, extraversion-introversion and stability-neuroticism, on 3570 individuals collected in the early 1990s. 

The summary of a varrank analysis on the Longley dataset is displayed below. One has to choose a model and a discretization method. The output is a list with two entries: the ordered names of the candidate variables and the triangular matrix of scores. For example, the variable \textit{glucose} is chosen first because the MI is the largest amongst all the variables (0.187). Then the variable \textit{mass} is chosen because the score 0.04 is the highest score when variables \textit{diabetes} and \textit{glucose} are already selected. At this step the variable \textit{insulin} has a negative score ($-0.041$), indicating that the relevant part of the score is smaller than the redundant part of the score. 

\begin{Verbatim}[fontsize=\footnotesize]
> install.packages("varrank")
> library(varrank)
> summary(varrank(data.df = PimaIndiansDiabetes,
+ 	method = "esteves",
+ 	variable.important = "diabetes",
+ 	discretization.method = "sturges",
+ 	algo = "forward", scheme = "mid", verbose=FALSE))

Number of variables ranked: 8
forward search using esteves method 
(mid scheme) 
 
Ordered variables (decreasing importance):
       glucose mass   age pedigree insulin pregnant pressure triceps
Scores   0.187 0.04 0.036   -0.005  -0.013   -0.008   -0.014      NA

 ---
 
 Matrix of scores: 
         glucose   mass    age pedigree insulin pregnant pressure triceps
glucose    0.187                                                         
mass       0.092   0.04                                                  
age        0.085  0.021  0.036                                           
pedigree   0.031  0.007 -0.005   -0.005                                  
insulin    0.029 -0.041 -0.024   -0.015  -0.013                          
pregnant   0.044  0.013  0.017    -0.03  -0.016   -0.008                 
pressure   0.024 -0.009 -0.021   -0.024  -0.019   -0.015   -0.014        
triceps    0.034  0.009 -0.046   -0.034  -0.024   -0.035    -0.02  
\end{Verbatim}

\begin{figure}[!h]
\begin{subfigure}{.5\textwidth}
  \raggedleft
  \includegraphics[width=1\linewidth]{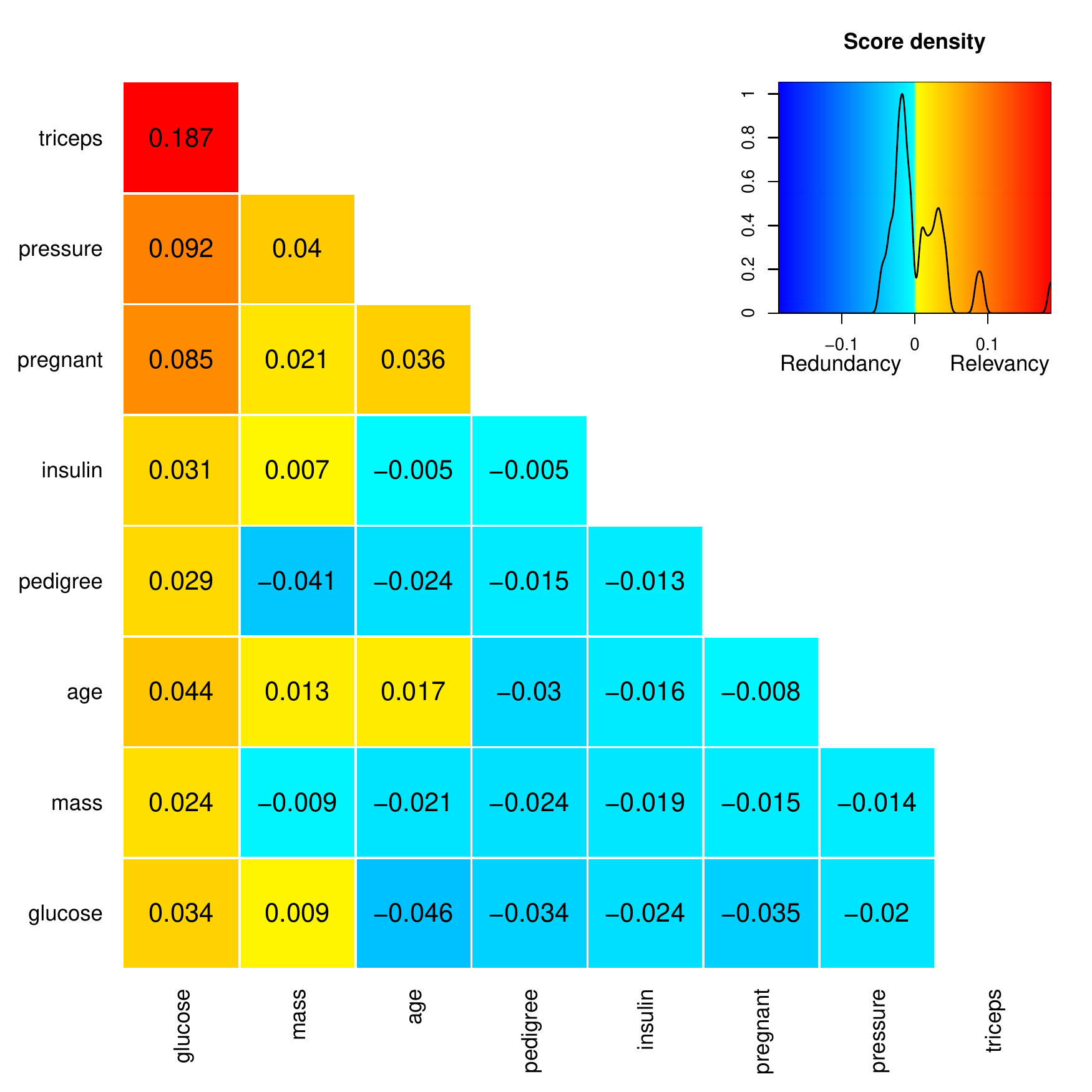}
  \caption{A: Pima Indians Diabetes}
  \label{fig:sfig1}
\end{subfigure}%
\begin{subfigure}{.5\textwidth}
  \centering
  \includegraphics[width=1\linewidth]{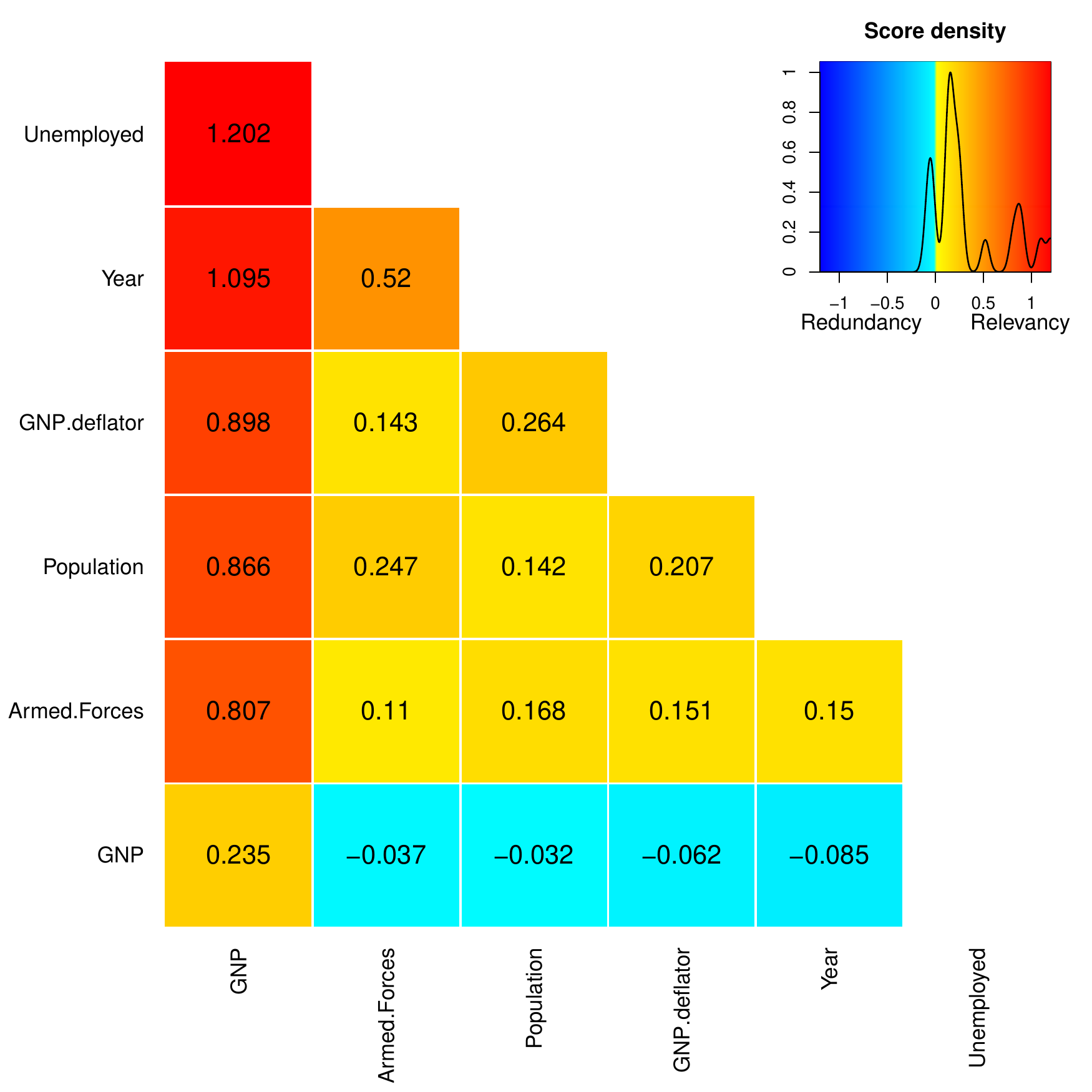}
  \caption{B: Longley}
  \label{fig:sfig2}
\end{subfigure}
\caption{Output of an analysis using {\sf varrank} for two datasets. The score matrix is displayed using both numerical values and color code. A key legend with the distribution of scores is also displayed.}
\label{fig:fig_Longley}
\end{figure}

Figure~\ref{fig:fig_Longley} (left panel) presents the analysis of the Pima Indians Diabetes dataset using {\sf varrank}. One can see a key legend with color coding (blue for redundancy, red for relevancy) and the distribution of the scores. The triangular matrix displays vertically the scores at each selection step. At each step, the variable with the highest score is selected (the variables are ordered in the plot). The scores at selection can be read from the diagonal. A negative score indicates a redundancy final trade of information and a positive score indicates a relevancy final trade of information. In the plot the scores are rounded to 3 digits. Figure~\ref{fig:fig_Longley} (right panel) presents the varrank analysis of the Longley dataset. 

\begin{figure}[!h]
  \centering
\includegraphics[width=0.8\linewidth]{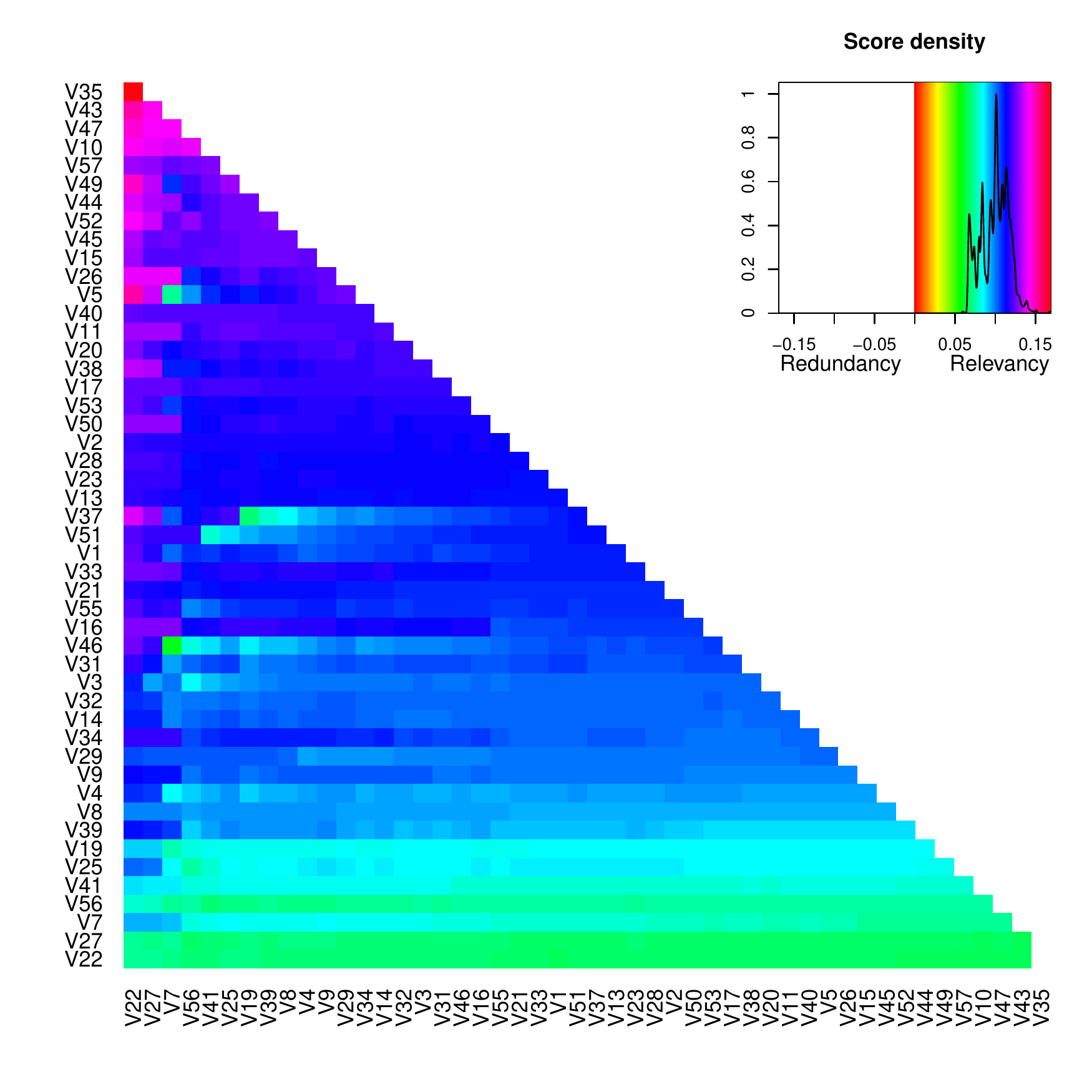}
  \caption{EPI dataset.}
  \label{fig:epi}
\end{figure}

The S3 plot function available in {\sf varrank} is flexible and allows the user to tailor the output, see Figure~\ref{fig:epi}. The final rendering depends on the algorithm used (see Figure~\ref{fig:PID_back} for an example from the backward algorithm). Additionally, a unique feature of {\sf varrank} is that it can deal with a set of variables of importance provided by a list of variable names, as one can see below.

\begin{Verbatim}[fontsize=\footnotesize]
> epi.varrank <- varrank(data.df = epi,
method = "peng",
variable.important = c("V6","V12","V18","V24","V30","V36","V42","V48","V54"),
discretization.method = "sturges",
algorithm = "forward",
scheme = "mid",
verbose = FALSE)
\end{Verbatim}

\begin{table}[!h]
\centering
\footnotesize
 \begin{tabular}{llll}
\multicolumn{4}{c}{}      \\ 
                     & {\sf varrank}   & {\sf caret}    & {\sf Boruta}   \\ \hline
\#1                  & glucose  &  glucose  & glucose  \\
\#2                  & mass     &  mass     & mass     \\
\#3                  & age      &  age      & age      \\
\#4                  & pedigree &  pregnant & pregnant \\
\#5                  & insulin  &  pedigree & insulin  \\
\#6                  & pregnant &  pressure & pedigree \\
\#7                  & pressure &  triceps  & triceps  \\
\#8                  & triceps  &  insulin  & pressure \\\hline
Bootstrapping 80\% & 29\% & 24\%  & 17\% \\ \hline
Running time {[}s{]} & 2.72 &  4.31     & 31.22    \\ \hline
\end{tabular}
\caption{Variable ranking comparison between {\sf varrank}, {\sf caret} and {\sf Boruta} for the Pima Indians Diabetes dataset.}
     \label{tab:stab1}
\end{table}

\begin{table}[!h]
\centering
\footnotesize
\begin{tabular}{clll}
\multicolumn{4}{c}{ }                                   \\ 
                     & {\sf varrank}   & {\sf caret}    & {\sf Boruta}       \\ \hline
\#1                  & GNP          & GNP          & GNP          \\
\#2                  & Armed.Forces & GNP.deflator & Year         \\
\#3                  & Population   & Year         & GNP.deflator \\
\#4                  & GNP.deflator & Population   & Population   \\
\#5                  & Year         & Armed.Forces & Armed.Forces \\
\#6                  & Unemployed   & Unemployed   & Unemployed   \\
\hline
Bootstrapping 80\% & 15\% & 0\%  & 0\% \\ \hline
Running time {[}s{]} & 0.07         & 0.89         & 0.57        \\ \hline
\end{tabular}
\caption{Variable ranking comparison between {\sf varrank}, {\sf caret} and {\sf Boruta} for the Longley dataset.}
\label{tab:stab2}
\end{table}

Tables \ref{tab:stab1} and \ref{tab:stab2} compare results from the packages {\sf varrank}, {\sf caret} and {\sf Boruta} for the Longley and Pima Indians Diabetes datasets. In {\sf varrank}, the esteves model, the most complex (thus the slowest) of the four possible models, was used with Sturges' rule as the discretization method in a forward search. In {\sf caret}, we used a learning vector quantization (lvq) and linear model (lm) for the classification and regression models for the Pima Indians Diabetes and Longley datasets, respectively. We used default settings for {\sf Boruta}.
The exact order of some of the less important variables depends on the package and method used. As one can see in Tables~\ref{tab:stab1} and~\ref{tab:stab2}, the Pima Indians Diabetes dataset exhibits a considerable degree of concordance. But the Longley dataset, which is known to be highly collinear, shows somewhat less agreement between the three approaches. The position of the variable `Armed.Forces' is quite different with {\sf varrank} as compared to the other methods. One important aspect in practice is the stability of the ranking. The 80\% bootstrapping tends to measure the variation of the ranked variables as a function of the sample size. To compute it, 100 datasets have been created with a sampling of 80\% of the data without replacement. The different approaches have been applied to those subsamples. Then the percentage of times the subsample ranked variable lists matched the original output is computed and presented as Bootstrapping 80\% in Tables \ref{tab:stab1} and \ref{tab:stab2}. The Pima Diabetes dataset has a retrieval rate of about a quarter based on 614 sampled observations. The low retrieval rate for the Langley dataset is explained by the 13 sampled observations.  Additionally, {\sf varrank} is computationally competitive in terms of benchmarking for small to medium size datasets.

\begin{figure}[!h]
  \centering
\includegraphics[width=1\linewidth]{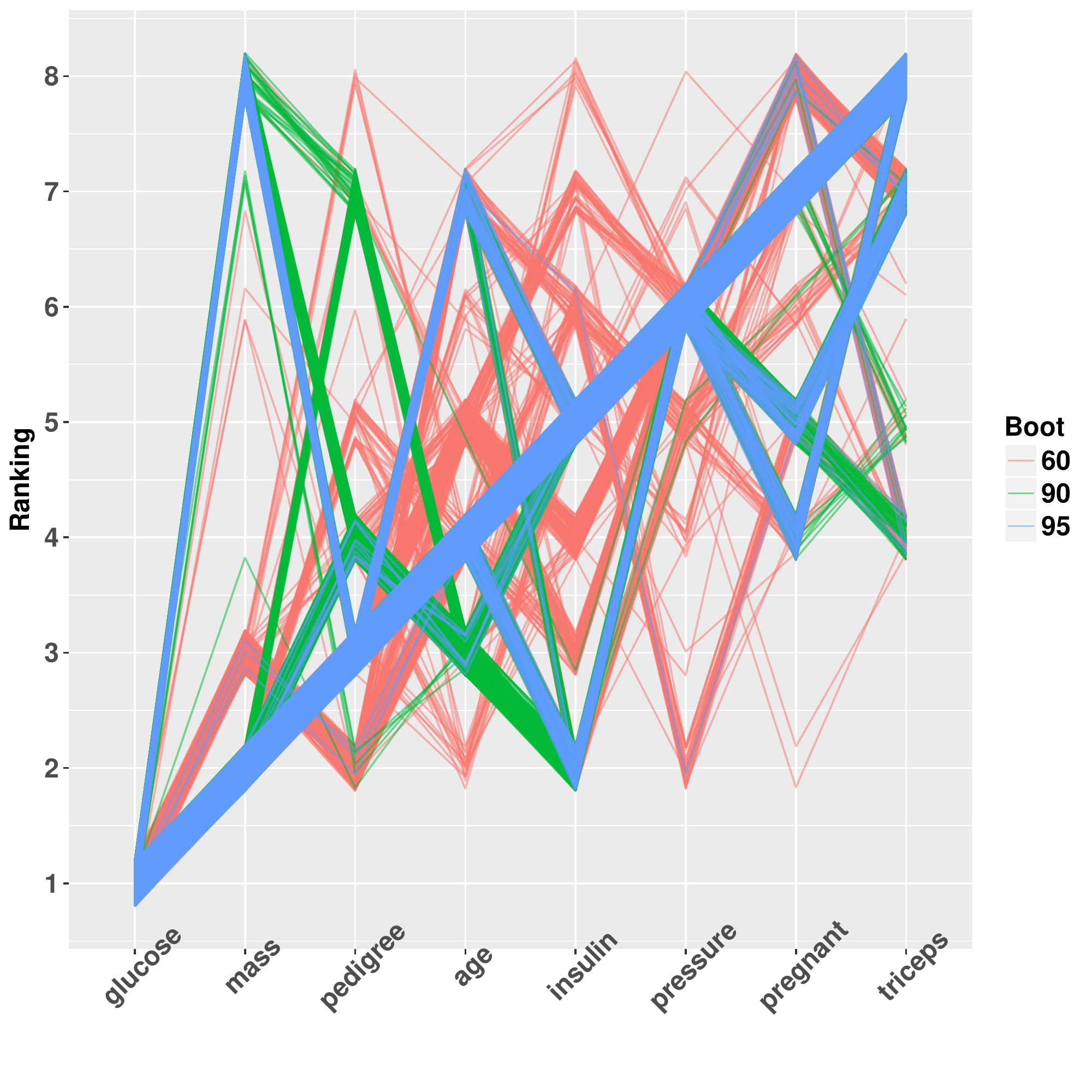}
  \caption{Parallel coordinates plot displaying the rank of the variables as a function of the bootstrapping sampling percentage.}
  \label{fig:percentage}
\end{figure}

Another measure of stability with sample size is presented in Figure~\ref{fig:percentage}. For each random sampling level, 1000 datasets were generated without replacement. Then a varrank analysis was performed. On the x-axis the order obtained with the full dataset is presented. On the y-axis, the retrieved rank is plotted. Each bootstrap sample leads to a trajectory in the graph. As one can see, the diagonal is quite visible, indicating that the variable ranking is confirmed by the bootstrap procedure. The global retrieved rank seems to increase with sample size. It also seems that some ranks have less uncertainty than others. The variable \textit{glucose} seems to have a high relevance for the variable \textit{diabetes} independent of the sample size chosen. The variable \textit{pressure}, on the other hand, often ranks sixth for 95\% and 90\% bootstrap random sampling levels. This suggests that the variables \textit{mass}, \textit{pedigree}, \textit{age} and \textit{insulin} are more relevant than \textit{pressure}. But their relative rank is not totally determined. 
  
\section{Conclusion and impact}

Many epidemiologists have expressed the need for a flexible, model-free implementation of a variable selection algorithm. One typical candidate for multivariable selection approach is Bayesian network modeling, which often uses multiple variables as target sets \cite{lewis2011structure}. Indeed, it often requires preselection of the candidate variable(s) for computational reasons \cite{lewis2013improving}. Generally, in machine learning approaches, integrating many or all possible variables in the analysis will lead to a slowdown and a decrease in accuracy of the inference process. Traditionally, the main approach to variable selection in epidemiology is prior knowledge from the scientific literature \cite{walter2009variable}. Then secondly, approaches with a pre-specified change-in estimate criterion and stepwise model selection are together as important as prior knowledge approach for variable selection \cite{walter2009variable}. However, these approaches have been disparaged for requiring arbitrary thresholds that can lead to biased estimates or overfitting of the true effect \cite{greenland2008invited, mickey1989impact}. 

Those approaches struggle to scale with problem dimensionality. Other multivariate machine learning approaches, such as principal components analysis or penalized regression, are rarely used. The latter approach assumes one single outcome. This is not always suitable in systems epidemiology where the focus could be more on the dynamics and relationship understanding between variables instead of predicting a given outcome. Existing approaches often rely on the assumption that predictive power is a proxy for importance in the modeling procedure. This is certainly reasonable if the main interest is to make predictions, but systems epidemiology is concerned with the underlying structure and prediction is viewed as a consequence. This is why approaches relying on mRMRe are conceptually seductive as they tends to optimize association penalized with redundancy. Then a flexible and model-free approach implemented in R that can be used jointly with expert knowledge for variable selection could help to better allocate time for variable selection. {\sf varrank} has been developed especially for this purpose: (i) it has visual output that is designed to help in the data analysis, (ii) it can handle multiple variables of importance and thus is a multivariable approach that goes beyond the classical paradigm of the one-outcome framework and (iii) it has a wide class of models implemented, with many options configurable by the end user.




\clearpage

\bibliographystyle{elsarticle-num-names}
\bibliography{bib_varrank}


%
%
%
%
%

\section*{Required Metadata}

\section*{Current code version}

\begin{table}[!h]
\begin{tabular}{|l|p{6.5cm}|p{6.5cm}|}
\hline
\textbf{Nr.} & \textbf{Code metadata description} & \textbf{Please fill in this column} \\
\hline
C1 & Current code version & v0.1 \\
\hline
C2 & Permanent link to code/repository used for this code version & \url{https://git.math.uzh.ch/gkratz/varrank} \\
\hline
C3 & Legal Code License   & GPL-2 \\
\hline
C4 & Code versioning system used & git \\
\hline
C5 & Software code languages, tools, and services used & R 3.4.0 or later from \url{https://cran.r-project.org/} \\
\hline
C6 & Compilation requirements, operating environments \& dependencies & Linux, OS X, Microsoft Windows. Runs within the R software environment\\
\hline
C7 & If available Link to developer documentation/manual & \url{https://git.math.uzh.ch/gkratz/varrank/varrank.pdf} \\
\hline
C8 & Support email for questions & gilles.kratzer@math.uzh.ch \\
\hline
\end{tabular}
\caption{Code metadata.}
\label{}
\end{table}

\section*{Current executable software version}
\label{}

\begin{table}[!h]
\begin{tabular}{|l|p{6.5cm}|p{6.5cm}|}
\hline
\textbf{Nr.} & \textbf{(Executable) software metadata description} & \textbf{Please fill in this column} \\
\hline
S1 & Current software version & v0.1 \\
\hline
S2 & Permanent link to executable of this version  & \url{https://git.math.uzh.ch/gkratz/varrank} \\
\hline
S3 & Legal Software License & GPL-2 \\
\hline
S4 & Computing platforms/Operating Systems & Linux, OS X, Microsoft Windows. Runs within the R software environment \\
\hline
S5 & Installation requirements \& dependencies & R 3.4.0 or later from https://cran.r-project.org/\\
\hline
S6 & If available, link to user manual - if formally published include a reference to the publication in the reference list & \url{https://git.math.uzh.ch/gkratz/}varrank/varrank.pdf \\
\hline
S7 & Support email for questions & gilles.kratzer@math.uzh.ch\\
\hline
\end{tabular}
\caption{Software metadata.}
\label{}
\end{table}

\clearpage

\section*{Supplementary material}\label{appendix:algo}

The sequential forward and backward algorithms are described using pseudocode \ref{algo:mRMR}. Let $\textbf{C}$ be the set of variables of importance and let $\textbf{F}$ be the set of  variables to rank. We assume   $\textbf{C}$ and $\textbf{F}$ are disjoint and let  $\textbf{D} = \textbf{C} \cup \textbf{F}$, i.e.,  $\textbf{D}$ is the data matrix consisting of observations (rows) of the variables (columns).

\SetArgSty{textrm}

\subsection{Forward and backward algorithms}

\begin{algorithm}[!h]
 \KwData{A dataset $\textbf{D}$, such that $\textbf{D} = \textbf{C} \cup \textbf{F}$, where  $\textbf{C}$ is the set of variables of importance and $\textbf{F}$ is the set of variables to rank.}
 \KwResult{$\textbf{S}$, the ranked set of variables $\textbf{F}$}\bigskip
 Initialization\;
 Set $\textbf{S} \leftarrow \varnothing$\;
$\textbf{S} \leftarrow f_i$ where $f_i$ = $\argmax_{f \in \textbf{F}} \text{MI}(f;\textbf{C})$\;
$\textbf{F} \leftarrow \textbf{F} \setminus f_i$\;
 \While{$|  \textbf{S} | \leq | \textbf{F} |-1$}{
 
  $f_i = \argmax_{f \in \textbf{F}} g(\alpha, \beta,\textbf{C}, \textbf{S}, f) = \text{MI}(f_i;\textbf{C}) - \sum_{f_s \in \textbf{S}} \alpha( \beta, f_i,f_s,\textbf{C}, \textbf{S}) \text{MI}(f_i;f_s)$\;

   $\textbf{S} \leftarrow \textbf{S} \cup f_i$\;
   
  $\textbf{F} \leftarrow \textbf{F}\setminus f_i $
   }
   $\textbf{S} \leftarrow \textbf{S} \cup \textbf{F}$
\bigskip

 \caption{The forward mRMRe ranking algorithm with mid scheme.}\label{algo:mRMR}
\end{algorithm}

%

The backward algorithm prunes the full set $\textbf{F}$ by minimizing the mRMRe equation \eqref{eq:mRMR}. The very first variable is chosen according to the scoring equation and not purely based on mutual information. The rest of the algorithm is mostly unchanged.

\subsection{Backward display}

\begin{figure}[!h]
  \centering
\includegraphics[width=0.8\linewidth]{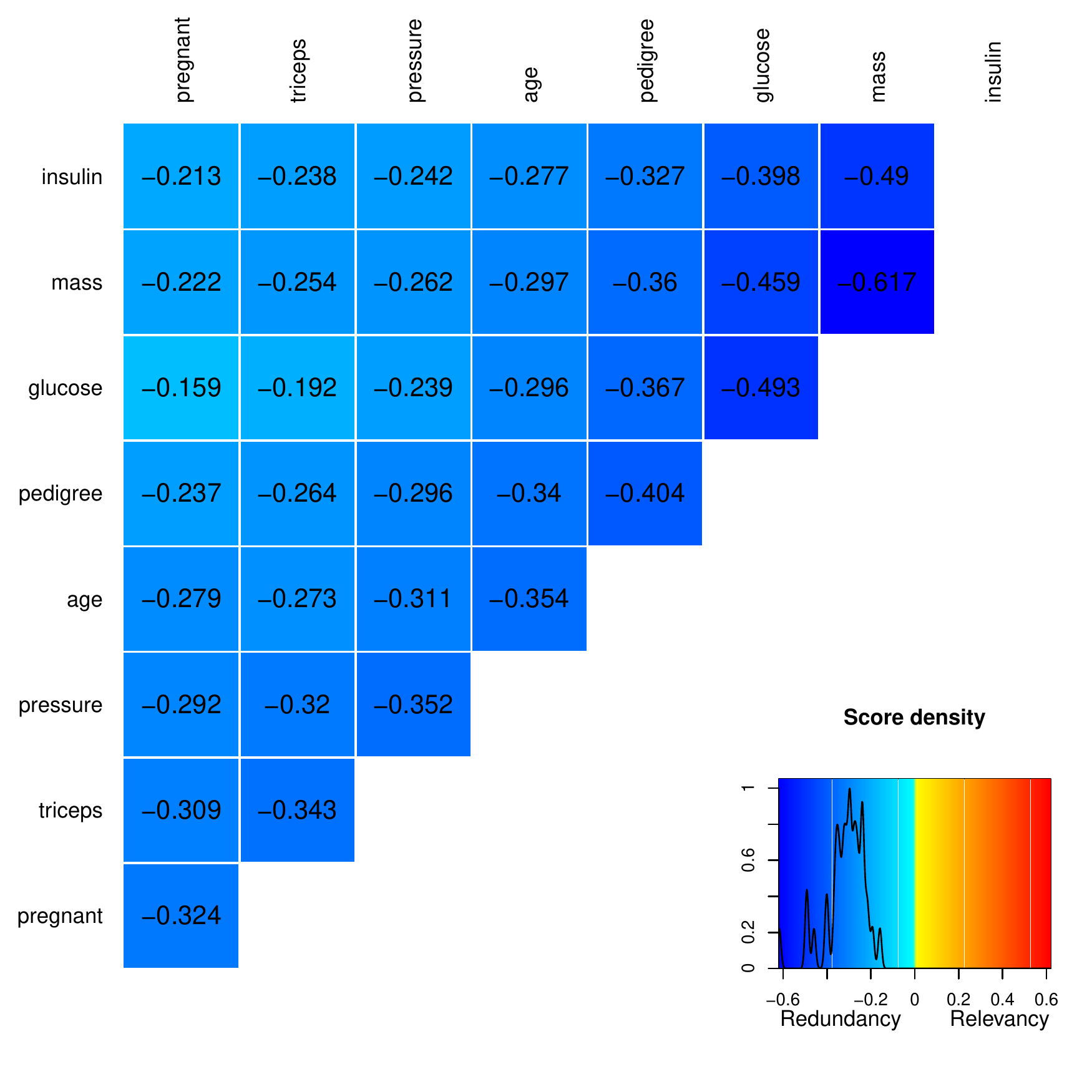}
  \caption{A backward analysis of the Pima Indians Diabetes dataset with mid scheme.}
  \label{fig:PID_back}
\end{figure}

Figure \ref{fig:PID_back} shows an example of a plot from a backward \textit{peng} search using a \textit{mid} scheme on the Pima Indians Diabetes dataset. As one can see, the triangular matrix is plotted back to front to highlight the difference between the backward and forward searches.

%
%

\end{document}